\documentclass[11pt]{article}
\usepackage{acl2015}
\usepackage{times}
\usepackage{url}
\usepackage{latexsym}
\usepackage{epsfig}
\usepackage{array}
\usepackage{color}
\usepackage{multirow}
\usepackage{url}
\usepackage{subfig}
\usepackage{algorithmic}
\usepackage{algorithm}
\usepackage{multirow}
\usepackage{rotating}
\usepackage{booktabs}
\usepackage{makecell}
\usepackage{comment}
\usepackage{amsmath,amsfonts,amssymb}

\newcommand*{\Scale}[2][4]{\scalebox{#1}{$#2$}}%

\title{Stochastic Language Generation in Dialogue using Recurrent Neural Networks with Convolutional Sentence Reranking}

\author{Tsung-Hsien Wen, Milica Ga{\v{s}}i\'c, Dongho Kim, Nikola Mrk{\v{s}}i\'c, \\ {\bf Pei-Hao Su, David Vandyke \and  Steve Young} \\
  Cambridge University Engineering Department,  \\
  Trumpington Street, 
  Cambridge, CB2 1PZ, UK\\
  {\tt \{thw28,mg436,dk449,nm480,phs26,djv27,sjy\}@cam.ac.uk}\
 }

\date{}
\begin{document}
\maketitle

\begin{abstract}
The natural language generation (NLG) component of a spoken dialogue system (SDS) usually needs a substantial amount of handcrafting or  a well-labeled dataset to be trained on.
These limitations add significantly to development costs and make  cross-domain, multi-lingual dialogue systems intractable. 
Moreover, human languages are context-aware. The most natural response should be directly learned from data rather than depending on predefined syntaxes or rules.
This paper presents a statistical language generator based on a joint recurrent and convolutional neural network structure which can be  trained on  dialogue act-utterance pairs without any semantic alignments or predefined grammar trees.
Objective metrics suggest that this new model outperforms previous methods under the same experimental conditions.
Results of an evaluation by human judges indicate that it produces not only high quality but linguistically varied utterances which are preferred compared to n-gram and rule-based systems.
\end{abstract}

\section{Introduction}
\label{sec:intro}

Conventional spoken dialogue systems (SDS) are expensive to build  because many of the processing components require a substantial amount of handcrafting \cite{Ward1994,Bohus2009}.
In the past decade, significant progress has been made in applying statistical methods to automate the speech understanding and dialogue management components of an SDS, including making them more easily extensible to other application domains \cite{6407655,gavsic2014incremental,Henderson2014d}.
However, due to the difficulty of collecting semantically-annotated corpora, the use of data-driven NLG for SDS remains relatively unexplored and rule-based generation remains the norm for most systems \cite{cheyer2007method,mirkovic2011dialogue}.

The goal of the NLG component of an SDS is to map an abstract dialogue act consisting of an act type and a set of attribute-value pairs\footnote{
Here and elsewhere, attributes are frequently referred to as {\it slots}.} into an appropriate surface text (see Table~\ref{tab:example} below for some examples).  An early example of a statistical NLG system is HALOGEN by \newcite{Langkilde1998} which uses an n-gram language model (LM) to rerank a set of candidates generated by a handcrafted generator.
In order to reduce the amount of handcrafting and make the approach more useful in SDS, \newcite{Oh2000} replaced the handcrafted generator with a set of word-based n-gram LM-based generators, one for each dialogue type and then reranked the generator outputs using a set of rules to produce the final response.
Although \newcite{Oh2000}'s approach 
limits the amount of handcrafting to a small set of post-processing rules, their system incurs a large computational cost in the over-generation phase and it is difficult to ensure that all of the required semantics are covered by the selected output.
More recently,  a phrase-based NLG system called BAGEL trained from utterances aligned with coarse-grained semantic concepts has been described  \cite{Mairesse2010,Mairesse2014}.  By implicitly modelling paraphrases, Bagel can  generate linguistically varied utterances.
However, collecting semantically-aligned corpora is expensive and time consuming, which limits Bagel's scalability to new domains.

This paper presents a neural network based NLG system that can be fully trained from dialog act-utterance pairs without any semantic alignments between the two.
We start in Section \ref{sec:rgm} by presenting a generator based on a recurrent neural network language model (RNNLM) \cite{39298195,5947611} which is trained on a {\it delexicalised} corpus \cite{Henderson2014d} 
whereby each value has been replaced by a symbol representing its corresponding slot.  In a final post-processing phase, these slot symbols are converted back to the corresponding slot values.

While generating, the RNN generator is conditioned on an auxiliary dialogue act feature and a controlling gate to over-generate candidate utterances for subsequent reranking.
In order to account for arbitrary slot-value pairs that cannot be routinely delexicalized in our corpus, Section \ref{sec:cnn} describes a convolutional neural network (CNN) \cite{Collobert2008,KalchbrennerGB14} sentence model which is used to validate the semantic consistency of candidate utterances during reranking.
Finally, by adding a backward RNNLM
reranker into the model in Section \ref{sec:brnn}, output fluency is further improved. 
Training and decoding details of the proposed system are described in Section \ref{sec:train} and \ref{sec:decode}.

Section \ref{sec:exp} presents an evaluation of the proposed system in the context of an application providing information about restaurants in the San Francisco area.  
In Section \ref{sec:compare}, we first show that new generator outperforms \newcite{Oh2000}'s utterance class LM approach using objective metrics, whilst at the same time being more computationally efficient. 
In order to assess the subjective performance of our system, pairwise preference tests are presented in Section \ref{sec:human_eval}. 
The results show that our approach can produce high quality utterances that are considered to be more natural than a rule-based generator.
Moreover, by sampling utterances from the top reranked output, our system can also generate linguistically varied utterances.
Section \ref{sec:analysis} provides a more detailed analysis of the contribution of each component of the system to the final performance.
We conclude with a brief summary and future work in Section \ref{sec:conclusion}.

\section{Related Work}
\label{sec:related} 

Conventional approaches to NLG typically divide the task into sentence planning, and surface realisation.
Sentence planning maps input semantic symbols into an intermediary tree-like or template structure representing the utterance, then  surface realisation converts the intermediate structure into the final text \cite{walker2002training,Stent04trainablesentence,Dethlefs13conditionalrandom}.  As noted above, one of the first 
statistical NLG methods that requires almost no handcrafting or semantic alignments was an n-gram based approach by \newcite{Oh2000}.
\newcite{Ratnaparkhi2002435} later addressed the limitations of n-gram LMs in the over-generation phase by using a more sophisticated generator based on a syntactic dependency tree.  

Statistical approaches have also been studied for sentence planning, for example, 
generating the most likely context-free derivations given a corpus \cite{Belz2008} or maximising the expected reward using reinforcement learning \cite{Rieser2010}. 
\newcite{Angeli2010} 
train a set of log-linear models to predict individual generation decisions given the previous ones, using only domain-independent features.
Along similar lines, by casting NLG as a template extraction and reranking problem, \newcite{kondadadi13} show that outputs produced by an SVM reranker are comparable to human-authored texts.

The use of neural network-based approaches to NLG
is relatively unexplored.
The stock reporter system ANA by \newcite{Kukich1987} is a network based NLG system, in which the generation task is divided into a sememe-to-morpheme network followed by a morpheme-to-phrase network.
Recent advances in recurrent neural network-based language models (RNNLM)  \cite{39298195,5947611} have demonstrated the value of distributed representations and the ability to model arbitrarily long dependencies for both speech recognition and machine translation tasks.
\newcite{ICML2011Sutskever_524} describes a simple variant of the RNN that can generate meaningful sentences 
by learning from a character-level corpus.
More recently, \newcite{KarpathyF14} have demonstrated that an RNNLM is capable of generating image descriptions by conditioning the network model on a pre-trained convolutional image feature representation.  This work provides a key inspiration for the system described here.
\newcite{zhangEMNLP2014} describes interesting work using RNNs to generate Chinese poetry.

A specific requirement of NLG for dialogue systems is that the concepts encoded in the abstract system dialogue act must be conveyed accurately by the generated surface utterance, and simple unconstrained RNNLMs  which rely on embedding at the word level \cite{NIPS2013_5021,pennington2014} are rather poor at this.
As a consequence, new methods have been investigated to learn distributed representations for phrases and even sentences by training models using different structures \cite{Collobert2008,SocherR2013Recursive}.
Convolutional Neural Networks (CNNs) were first studied in computer vision  for object recognition \cite{726791}.
By stacking several convolutional-pooling layers followed by a fully connected feed-forward network, CNNs are claimed to be able to extract several levels of translational-invariant features that are useful in classification tasks.
The convolutional sentence model \cite{KalchbrennerGB14,kim2014} adopts the same methodology but collapses the two dimensional convolution and pooling process into a single dimension.
The resulting model is claimed to represent the state-of-the-art for many speech and NLP related tasks \cite{KalchbrennerGB14,sainath2013deep}.

\section{Recurrent Generation Model}
\label{sec:rgm}

\begin{figure}[h]
\vspace{-3 mm}
\centerline{\includegraphics[width=80mm]{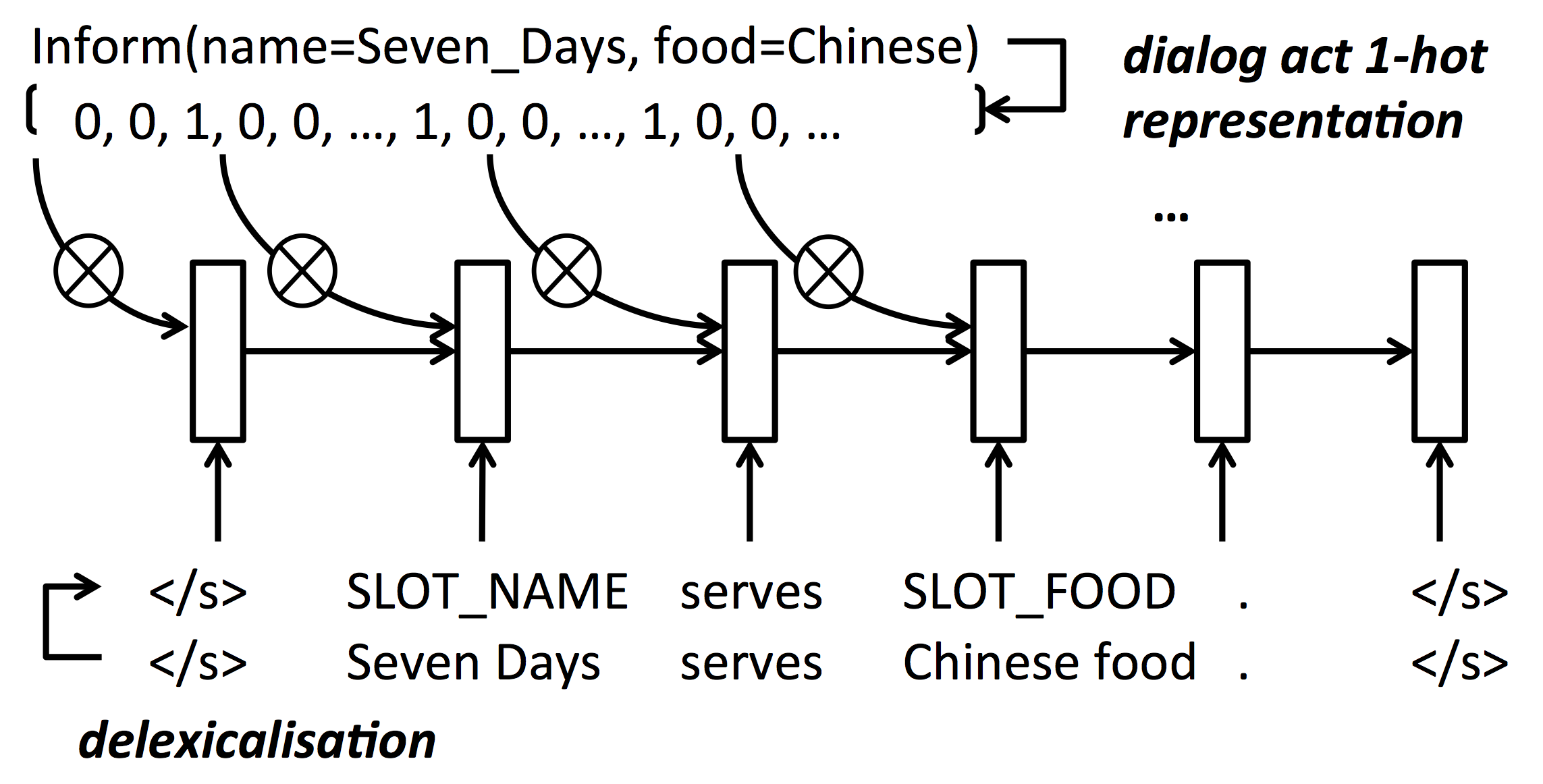}}
\caption{ An unrolled view of the RNN-based generation model. It operates on a delexicalised utterance and a 1-hot encoded feature vector specified by a dialogue act type and a set of slot-value pairs. $\otimes$ indicates the gate  used for controlling the on/off states of certain feature values.  The output connection layer is omitted here for simplicity.}
\label{fig:rgm}
\end{figure}

The generation model proposed in this paper is based on an RNNLM architecture \cite{39298195} in which a 1-hot encoding $\mathrm{\mathbf{w}}_{t}$ of a token\footnote{
We use {\it token} instead of {\it word} because our model operates on text for which slot names and values have been delexicalised.} 
$w_t$ is input at each time step $t$ conditioned on a recurrent hidden layer $\mathrm{\mathbf{h}}_t$
and outputs the probability distribution of the next token ${w}_{t+1}$.
Therefore, by sampling input tokens one by one from the output distribution of the RNN until a stop sign is generated \cite{KarpathyF14} or some required constraint is satisfied \cite{zhangEMNLP2014}, the network can produce a sequence of tokens which can be lexicalised to form
the required utterance.

In order to ensure that the generated utterance represents the intended meaning,  the input vectors $\mathrm{\mathbf{w}_t}$ are augmented by a control vector $\mathrm{\mathbf{f}}$
constructed from the concatenation of 1-hot encodings of the required dialogue act and its associated slot-value pairs.
The auxiliary information provided by this control vector tends to decay over time because of the {\it vanishing gradient problem} \cite{6424228,279181}.  Hence,  $\mathrm{\mathbf{f}}$ is reapplied to the RNN at every time step  as in \newcite{KarpathyF14}.

\begin{figure*}[t]
\centerline{\includegraphics[width=120mm]{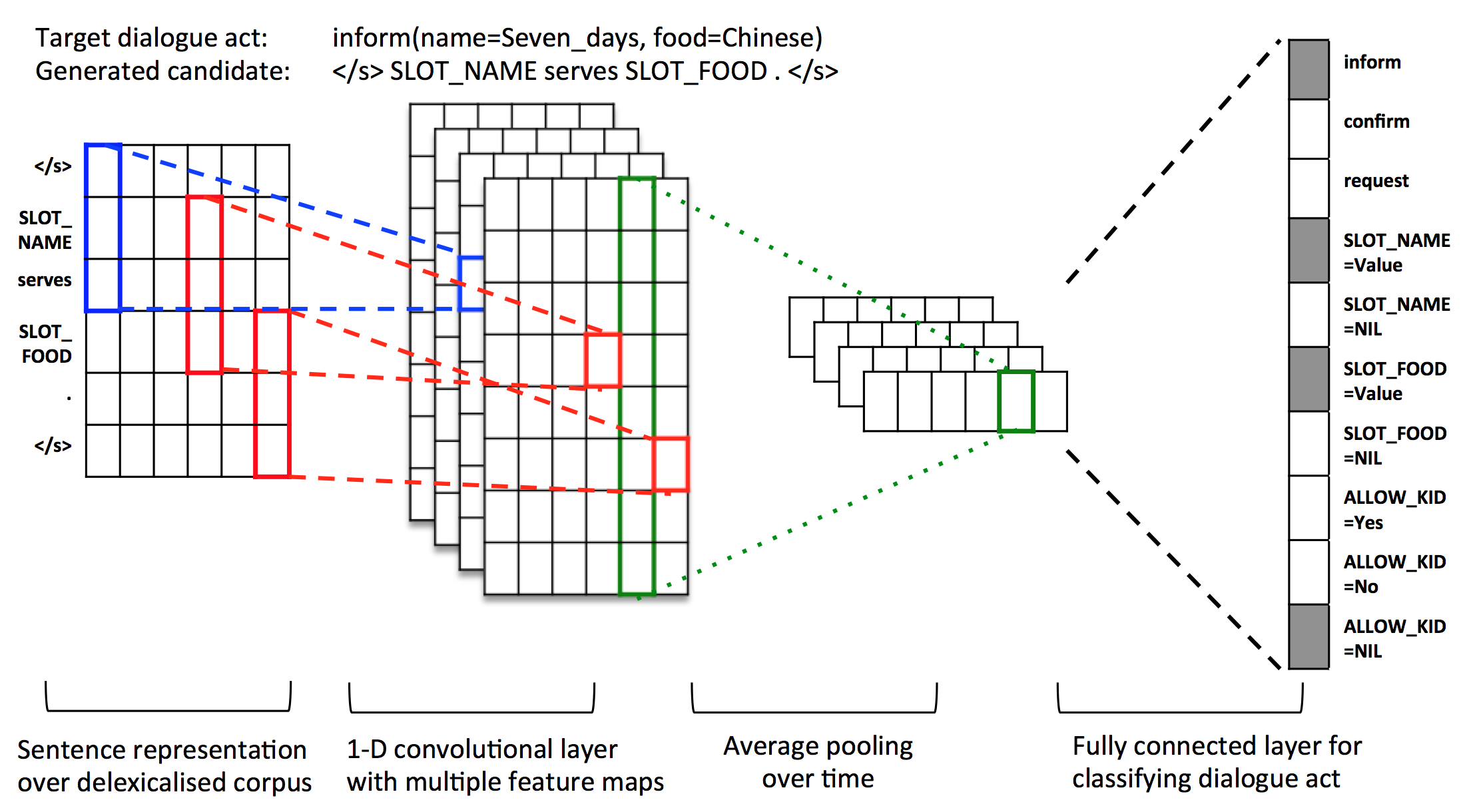}}
\caption{ Our simple variant of CNN sentence model as described in \newcite{KalchbrennerGB14}. }
\vspace{-2 mm}
\label{fig:cnn}
\end{figure*}

In detail, the recurrent generator shown in Figure \ref{fig:rgm} is defined as follows:
\begin{equation}
\Scale[0.85]{\mathrm{\mathbf{h}}_t = sigmoid( \mathrm{\mathbf{W}}_{hh}\mathrm{\mathbf{h}}_{t-1}  +  \mathrm{\mathbf{W}}_{wh}\mathrm{\mathbf{w}}_t + \mathrm{\mathbf{W}}_{fh}\mathrm{\mathbf{f}}_t)}
\end{equation}
\begin{equation}
\Scale[0.85]{P(w_{t+1}|w_t,w_{t-1},...w_0,\mathrm{\mathbf{f}}_t) = softmax( \mathrm{\mathbf{W}}_{ho}\mathrm{\mathbf{h}}_t )}
\end{equation}
\begin{equation}
\Scale[0.85]{w_{t+1} \sim P(w_{t+1}|w_t,w_{t-1},...w_0,\mathrm{\mathbf{f}}_{t})}
\end{equation}
where $\mathrm{\mathbf{W}}_{hh}$, $\mathrm{\mathbf{W}}_{wh}$, $\mathrm{\mathbf{W}}_{fh}$, and $\mathrm{\mathbf{W}}_{ho}$ are the learned network weight matrices.
$\mathrm{\mathbf{f}}_t$ is a gated version of $\mathrm{\mathbf{f}}$ designed to discourage duplication of information in the generated output in which
each segment $\mathrm{\mathbf{f}}_s$ of the control vector $\mathrm{\mathbf{f}}$ corresponding to slot $s$ is replaced by
\begin{equation}
\mathrm{\mathbf{f}}_{s,t} = \mathrm{\mathbf{f}}_s\odot\delta^{t-t_s}
\end{equation}
where $t_s$ is the time at which slot $s$ first appears in the output,
$\delta \leq1$ is a decay factor, and $\odot$ denotes element-wise multiplication.
The effect of this gating is to decrease the probability of regenerating slot symbols that have already been generated, and to increase the probability of rendering all of the information encoded in $\mathrm{\mathbf{f}}$.

The tokenisation resulting from delexicalising slots and values does not work for all cases.  For example, 
some slot-value pairs such as {\it food=dont\_care} or  {\it kids\_allowed=false} cannot be directly modelled using this technique because there is no explicit value to delexicalise in the training corpus.
As a consequence, the model is prone to errors when these slot-value pairs are required.
A further problem is that the RNNLM generator selects words based only on the preceding history, whereas some sentence forms depend on the backward context.

To deal with these issues, candidates generated by the RNNLM are reranked using two models.  Firstly, a convolutional 
neural network (CNN) sentence model \cite{KalchbrennerGB14,kim2014} is used to ensure that the required dialogue act and slot-value
pairs are represented in the generated utterance, including the non-standard cases.   Secondly, a  {\it backward} RNNLM is used
to rerank utterances presented in reverse order. 

\subsection{Convolutional Sentence Model}
\label{sec:cnn}


The CNN sentence model  is shown in Figure \ref{fig:cnn}. 
Given a candidate utterance of length $n$, an utterance matrix $\mathrm{\mathbf{U}}$ is constructed by stacking embeddings $\mathrm{w_t}$ of each token in the utterance:
\begin{equation}
\mathrm{\mathbf{U}} = \begin{bmatrix} 
\line(1,0){25} \ \ \mathrm{w_0} \ \ \line(1,0){25}\\
\line(1,0){25} \ \ \mathrm{w_1} \ \ \line(1,0){25}\\
... \\
\line(1,0){25} \ \ \mathrm{w_{n-1}} \ \ \line(1,0){25}
\end{bmatrix}.
\end{equation}
A set of $K$ convolutional mappings are then applied to the utterance to form a set of feature detectors.  The outputs of these detectors are combined and fed into a fully-connected feed-forward network to classify the action type and whether each required slot is mentioned or not. 

Each mapping  $k$ consists of a one-dimensional convolution  between a filter $\mathrm{\mathbf{m}_k}\in\mathbb{R}^m$ and the utterance matrix $\mathrm{\mathbf{U}}$ to produce another matrix $\mathrm{\mathbf{C}^k}$:
\begin{equation}
\mathrm{\mathbf{C}^k}_{i,j} = \mathrm{\mathbf{m}_k}^{\intercal}\mathrm{\mathbf{U}}_{i-m+1:i,j}
\end{equation}
where $m$ is the filter size, and $i$,$j$ is the row and column index respectively.
The outputs of each column of $\mathrm{\mathbf{C}^k}$ are then pooled by averaging\footnote{Max pooling was also tested but was found to be inferior to average pooling} over time:
\begin{equation}
\mathrm{\mathbf{h}_k} = \begin{bmatrix} 
\bar{\mathrm{\mathbf{C}}}^k_{:,0},\bar{\mathrm{\mathbf{C}}}^k_{:,1},...,\bar{\mathrm{\mathbf{C}}}^k_{:,h-1}
\end{bmatrix}
\end{equation}
where $h$ is the size of embedding and $k=1 \ldots K$. 
Last, the $K$ pooled feature vectors $\mathrm{\mathbf{h}}_k$ are passed through a nonlinearity function to obtain the final feature map. 

\subsection{Backward RNN reranking}
\label{sec:brnn}

As noted earlier, the quality of an RNN language model may be improved if both forward and backward contexts are considered. 
Previously, bidirectional RNNs \cite{schuster1997bidirectional} have been shown to be effective for handwriting recognition \cite{graves2008unconstrained}, speech recognition \cite{graves2013speech}, and machine translation \cite{D141003}.
However, applying a bidirectional RNN directly in our generator is not straightforward since the generation process is sequential in time. 
Hence instead of integrating the bidirectional information into a single unified network, the forward and backward contexts are utilised separately by firstly generating candidates using the forward RNN generator, then using the log-likelihood computed by a backward RNNLM to rerank the candidates. 

\subsection{Training}
\label{sec:train}

Overall the proposed generation architecture requires three models to be trained: a forward RNN generator, a CNN reranker, and a backward RNN reranker. 
The objective functions for training the two RNN models are the cross entropy errors between the predicted word distribution and the actual word distribution in the training corpus, whilst the objective for the CNN model is the cross entropy error between the predicted dialogue act and the actual dialogue act, summed over the act type and each slot.
An $l_2$ regularisation term is added to the objective function for every 10 training examples as suggested in \newcite{rnnlm_tool}.
The three networks share the same set of word embeddings,  initialised with pre-trained word vectors provided by \newcite{pennington2014}.
All costs and gradients are computed and stochastic gradient descent is used to optimise the parameters.
Both RNNs were trained with back propagation through time \cite{werbos1990backpropagation}.
In order to prevent overfitting, early stopping was implemented using a held-out validation set.

\subsection{Decoding}
\label{sec:decode}
\begin{table*}
\centering
\small
\caption{The 8 system dialogue acts with example realisations}
\vspace{-3 mm}
\setlength{\intextsep}{3pt plus 2pt minus 2pt} 
\setlength{\abovecaptionskip}{0pt}
\setlength{\belowcaptionskip}{0pt}
\label{tab:example}
\begin{tabular}{|l|l|}
\hline
{\bf\#}& 	{\bf Dialogue act and example realisations of our system, by sampling from top-5 candidates}\\
\hline
1	&	inform(name="stroganoff restaurant",pricerange=cheap,near="fishermans wharf")\\
	&	{\it stroganoff restaurant is a cheap restaurant near fishermans wharf .}\\
	& 	{\it stroganoff restaurant is in the cheap price range near fishermans wharf .}\\
\hline
2	& 	reject(kidsallowed=yes,food="basque")\\
	&	{\it unfortunately there are 0 restaurants that allow kids and serve basque .}\\
\hline
3	& 	informonly(name="bund shanghai restaurant", food="shanghainese")\\
	&	{\it i apologize , no other restaurant except bund shanghai restaurant that serves shanghainese .}\\
	& 	{\it sorry but there is no place other than the restaurant bund shanghai restaurant for shanghainese .}\\
\hline
4	& 	confirm(goodformeal=dontcare)\\
	&	{\it i am sorry . just to confirm . you are looking for a restaurant good for any meal ?}\\
	&	{\it can i confirm that you do not care about what meal they offer ?}\\
\hline
5	& 	request(near)\\
	&	{\it would you like to dine near a particular location ?}\\
\hline
6	& 	reqmore()\\
	&	{\it is there anything else i can do for you ?}\\
\hline
7	& 	select(kidsallowed=yes, kidsallowed=no)\\
	&	{\it are you looking for a restaurant that allows kids , or does not allow kids ?}\\
\hline
8	& 	goodbye()\\
	&	{\it thank you for calling . good bye .}\\
\hline
\end{tabular}
\end{table*}

The decoding procedure is split into two phases: (a) over-generation, and (b) reranking.
In the over-generation phase, the forward RNN generator conditioned on the given dialogue act, is used to sequentially generate utterances by 
random sampling of the predicted next word distributions.
In the reranking phase, the hamming loss $cost_{CNN}$ of each candidate is computed using the CNN sentence model 
and the log-likelihood $cost_{bRNN}$ is computed using the backward RNN.
Together with the  log-likelihood $cost_{fRNN}$ from the forward RNN, the reranking score $R$ is computed as:
\begin{equation}
\label{eq:score}
\Scale[0.9]{R = - ( cost_{fRNN} + cost_{bRNN} + cost_{CNN} )}.
\end{equation}
This is the reranking criterion used to  analyse each individual model in Section \ref{sec:analysis}.

Generation quality can be further improved  by introducing a slot error criterion $\mathrm{ERR}$, which is the {\it number of slots generated that is either redundant or missing}.  This is also used in \newcite{Oh2000}.
Adding this to equation (\ref{eq:score}) yields the final reranking score $R^*$:
\begin{equation}
\label{eq:score_star}
\begin{split}
R^* = - ( &cost_{fRNN} + cost_{bRNN} + \\
&cost_{CNN} +\lambda \mathrm{ERR} )
\end{split}
\end{equation}
In order to severely penalise nonsensical utterances, $\lambda$ is set to 100 for both the proposed RNN  system and our implementation of \newcite{Oh2000}'s n-gram based system.
This reranking criterion is used for both the automatic evaluation in Section \ref{sec:compare} and the human evaluation in Section \ref{sec:human_eval}.

\section{Experiments}
\label{sec:exp}

\subsection{Experimental Setup}
\label{sec:data}

The target application area for our generation system is a spoken
dialogue system providing information about restaurants in San Francisco. 
There are 8 system dialogue act types such as {\it inform} to present information about restaurants, {\it confirm} to check that
a slot value has been recognised correctly, and {\it reject} to advise that the user's constraints cannot be met (Table \ref{tab:example} gives the full list with examples);
and there are 12 attributes (slots): {\it name, count, food, near, price, pricerange, postcode, phone, address, area, goodformeal, and kidsallowed}, in which all slots are categorical except {\it kidsallowed} which is binary.

To form a training corpus, dialogues from a set of 3577 dialogues collected in a user trial of a statistical dialogue manager proposed by \newcite{6407655} were randomly sampled
and shown to workers recruited via the Amazon Mechanical Turk service. 
Workers were shown each dialogue turn by turn and asked to enter an appropriate system response in natural English corresponding to each system dialogue act.
The resulting corpus contains 5193  hand-crafted system utterances from 1006 randomly sampled dialogues.
Each categorical value was replaced by a token representing its slot, and slots that appeared multiple times in a dialogue act were merged into one. This resulted in 228 distinct dialogue acts.

The system was implemented using the Theano library \cite{bergstra2010,Theano2012}.
The system was trained by partitioning the 5193 utterances into a training set, validation set, and testing set in the ratio 3:1:1, respectively. 
The frequency of each action type and slot-value pair differs quite markedly across the corpus, hence up-sampling was used to make the corpus more uniform.
Since our generator works stochastically and the trained networks can differ depending on the initialisation, all the results shown below\footnote{Except human evaluation, in which only one set of network was used.} were averaged over 10 randomly initialised networks.
The BLEU-4 metric was used for the objective evaluation \cite{papineni2002bleu}.
Multiple references for each test dialogue act were obtained by mapping them back to the 228 distinct dialogue acts, merging those delexicalised templates that have the same dialogue act specification, and then lexicalising those templates back to form utterances.
In addition, the slot error (ERR) as described in Section \ref{sec:decode}, out of 1848 slots in 1039 testing examples, was computed alongside the BLEU score.

\subsection{Empirical Comparison}
\label{sec:compare}

\begin{table}[t]
\vspace{-3 mm}
\caption{Comparison of top-1 utterance between the RNN-based system and three baselines. A two-tailed Wilcoxon rank sum test was applied to compare the RNN model with the best O\&R system (the 3-slot, 5g configuration) over 10 random seeds. (*=p\textless.005)}
\vspace{-2 mm}
\centering 
\setlength{\intextsep}{3pt plus 2pt minus 2pt} 
\setlength{\abovecaptionskip}{0pt}
\setlength{\belowcaptionskip}{0pt}
\label{tab:compare}
\hspace*{0pt}\makebox[\linewidth][c]{%
\scalebox{1.1}{
\begin{tabular}{lccc}
\Xhline{2\arrayrulewidth}
{\bf Method}	&	{\bf beam}	&	{\bf BLEU}	&	{\bf ERR}	\\
\Xhline{2\arrayrulewidth}
handcrafted	&	n/a		&	0.440	&	0		\\
\hline
kNN			&	n/a		&	0.591	&	17.2		\\
\hline
O\&R,0-slot,5g	&	1/20		&	0.527	& 	635.2	\\
O\&R,1-slot,5g	&	1/20		&	0.610	& 	460.8	\\
O\&R,2-slot,5g	&	1/20		&	0.719	& 	142.0	\\
\hline
O\&R,3-slot,3g	&	1/20		&	0.760	& 	74.4		\\
O\&R,3-slot,4g	&	1/20		&	0.758	& 	53.2		\\
O\&R,3-slot,5g	&	1/20		&	0.757	& 	47.8		\\
\hline
Our Model		&	1/20		&	{\bf 0.777*}& 	{\bf 0*}	\\	
\Xhline{2\arrayrulewidth}
\end{tabular}
}}
\end{table}
As can be seen in Table \ref{tab:compare}, we compare our proposed RNN-based method with three baselines: a handcrafted generator, a k-nearest neighbour method (kNN), and \newcite{Oh2000}'s n-gram based approach (O\&R).
The handcrafted generator was tuned over a long period of time and has been used frequently to interact with real users. We found its performance is reliable and robust.
The kNN was performed by computing the similarity of the testing dialogue act 1-hot vector against all training examples.
The most similar template in the training set was then selected and lexicalised as the testing realisation.
We found our RNN generator significantly outperforms these two approaches.
While comparing with the O\&R system, we found that by partitioning the corpus into more and more utterance classes, the O\&R system can also reach a BLEU score of 0.76.
However, the slot error cannot be efficiently reduced to zero even when using the error itself as a reranking criterion. This problem is also noted in \newcite{Mairesse2014}.
\begin{figure}[t]
\vspace{-3 mm}
\centerline{\includegraphics[width=82mm]{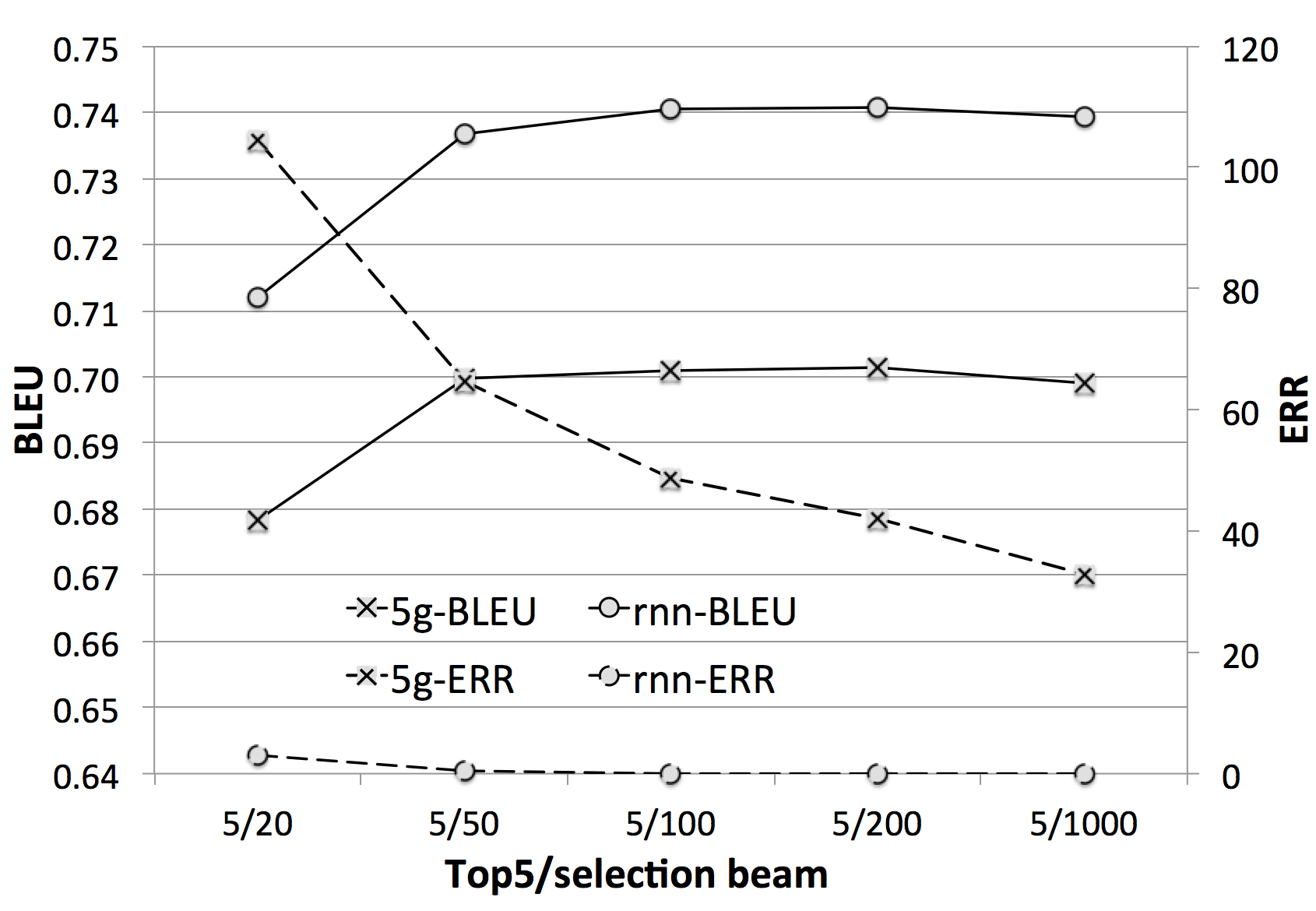}}
\caption{Comparison of our method (rnn) with O\&R's approach (5g) in terms of optimising top-5 results over different selection beams.}
\vspace{-4 mm}
\label{fig:compare_top5}
\end{figure}
\begin{table*}
\caption{Pairwise comparison between four systems. Two quality evaluations (rating out of 5) and one preference test were performed in each case. Statistical significance was computed using a two-tailed Wilcoxon rank sum test and a two-tailed binomial test (*=p\textless.05, **=p\textless.005). }
\centering 
\vspace{-4 mm}
\setlength{\intextsep}{3pt plus 2pt minus 2pt} 
\setlength{\abovecaptionskip}{0pt}
\setlength{\belowcaptionskip}{0pt}
\label{tab:human}
\hspace*{0pt}\makebox[\linewidth][c]{%
\scalebox{0.95}{
\begin{tabular}{c|cc||cc||cc||cc}
\hline
\multirow{2}{*}{Metrics}	&	handcrafted	&	RNN$_1$		&	handcrafted	&	RNN$_5$		&	RNN$_1$	&	RNN$_5$	&	O\&R$_5$ &	RNN$_5$ \\
\cline{2-9}
					&	\multicolumn{2}{c||}{148 dialogs, 829 utt.}	& 	\multicolumn{2}{c||}{148 dialogs, 814 utt.}	& 	\multicolumn{2}{c||}{144 dialogs, 799 utt.} &	\multicolumn{2}{c}{145 dialogs, 841 utt.}\\
\hline
Info.			&	3.75			&	3.81			&	3.85			&	3.93*			&	3.75		&	3.72		&	4.02		&	4.15*	\\
Nat.			&	3.58			&	3.74**		&	3.57			&	3.94**		&	3.67		&	3.58		&	3.91		&	4.02\\
\hline
Pref.			&	44.8\%		&	55.2\%*		&	37.2\%		&	62.8\%**		&	47.5\%	&	52.5\%	&	47.1\%	&	52.9\%\\
\hline
\end{tabular}
}}
\vspace{-3 mm}
\end{table*}

In contrast, the RNN system produces utterances without slot errors when reranking using the same number of candidates, and it achieves the highest BLEU score.
Figure \ref{fig:compare_top5} compares the RNN system with O\&R's system when randomly selecting from the top-5 ranked results in order to introduce linguistic diversity.
Results suggest that although O\&R's approach improves as the selection beam increases, the RNN-based system is still better in both metrics.
Furthermore, the slot error of the RNN system drops to zero when the selection beam is around 50. 
This indicates that the RNN system is capable of generating paraphrases by simply increasing the number of candidates during the over-generation phase.

\subsection{Human Evaluation}
\label{sec:human_eval}
\begin{figure*}
\minipage{0.32\textwidth}
  \includegraphics[width=\linewidth]{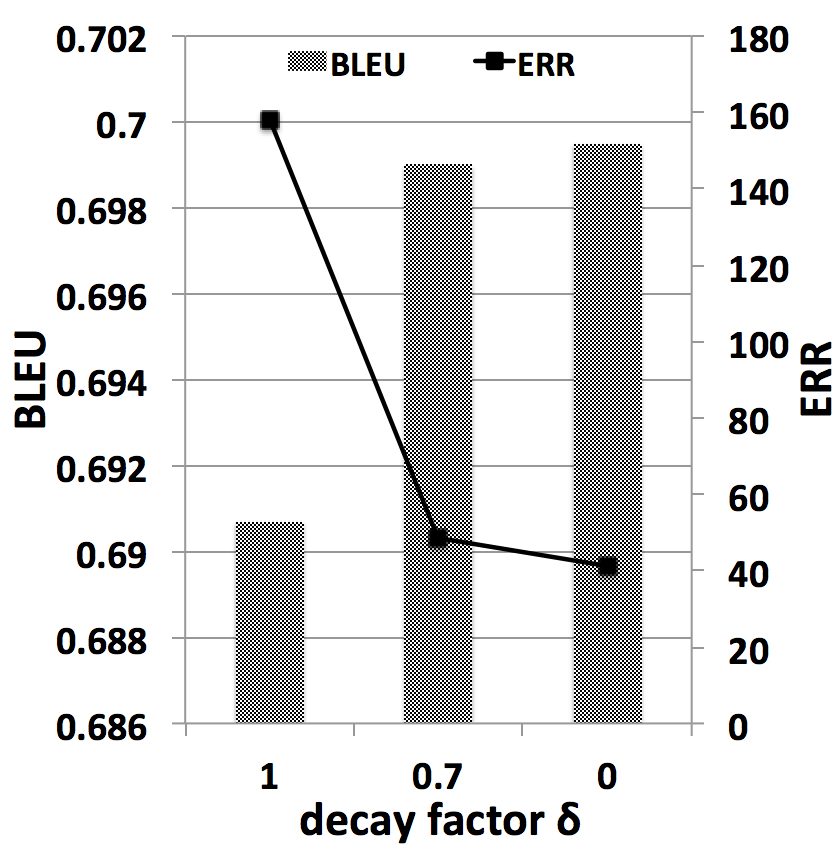}
  \caption{Feature gating effect}\label{fig:decay}
\endminipage\hfill
\minipage{0.32\textwidth}
  \includegraphics[width=\linewidth]{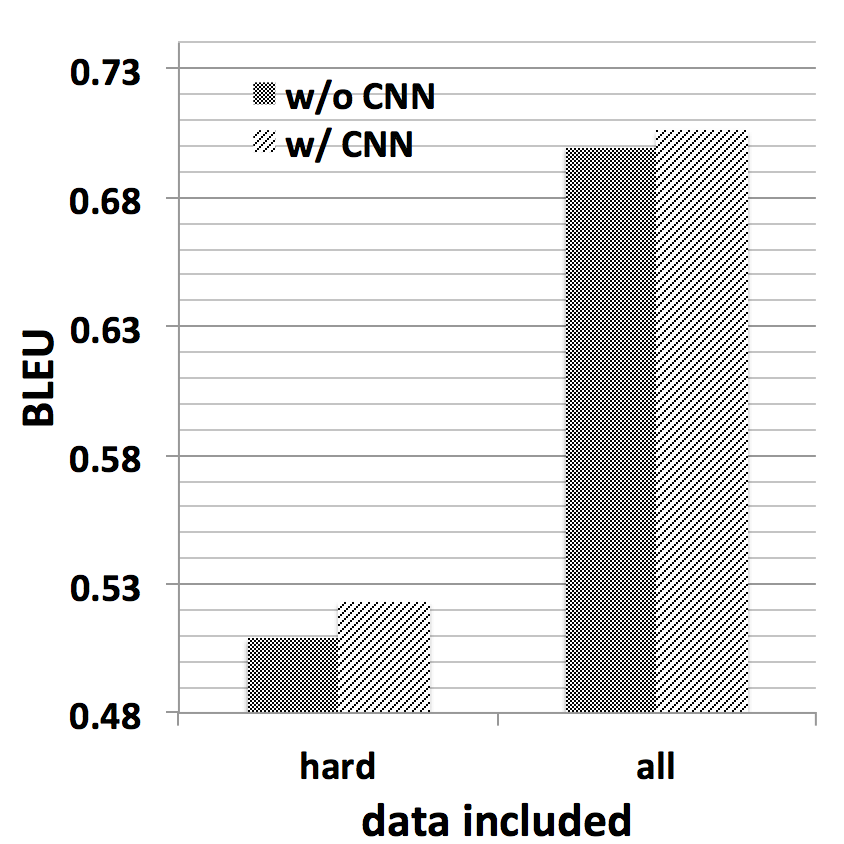}
  \caption{CNN effect}\label{fig:slu}
\endminipage\hfill
\minipage{0.32\textwidth}%
  \includegraphics[width=\linewidth]{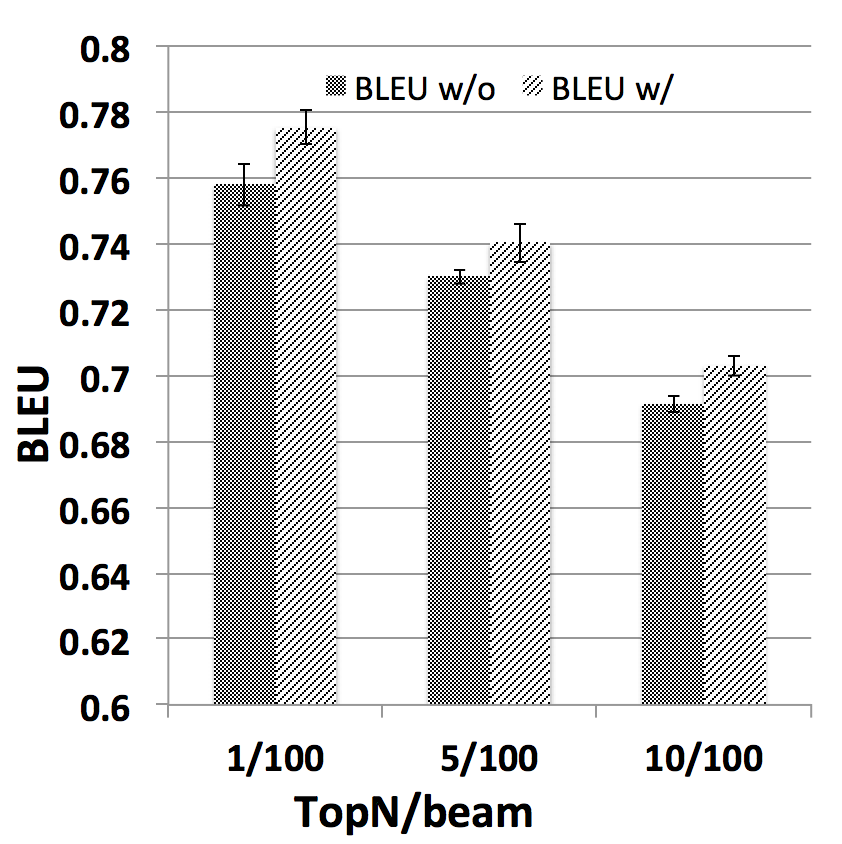}
  \caption{Backward RNN effect}\label{fig:sample}
\endminipage
\vspace{-4mm}
\end{figure*}

Whilst automated metrics provide useful information for comparing different systems, human testing is needed to assess subjective quality.
To do this, about 60 judges were recruited using Amazon Mechanical Turk and system responses were generated for the remaining 2571 unseen dialogues mentioned in Section \ref{sec:data}.   Each judge was then shown a randomly selected dialogue, turn by turn.
At each turn, two utterances were generated from two different systems and presented to the judge
who was asked to score each utterance in terms of informativeness and naturalness (rating out of 5), and also asked to state a preference between the two  taking account of the given dialogue act and the dialogue context.
Here {\it informativeness} is defined as whether the utterance contains all the information specified in the dialogue act, and {\it naturalness} is defined as whether the utterance could have been produced by a human.
The trial was run pairwise across four systems: the RNN system using 1-best utterance RNN$_1$, the RNN system sampling from the top 5 utterances RNN$_5$, the O\&R approach sampling from top 5 utterances O\&R$_5$, and a handcrafted baseline.

The result is shown in Table \ref{tab:human}.  As can be seen, 
the human judges preferred both RNN$_1$ and RNN$_5$ compared to the rule-based generator and the preference is statistically significant.
Furthermore, the RNN systems scored higher in both informativeness and naturalness metrics, though the difference for informativeness is not statistically significant. 
When comparing RNN$_1$ with RNN$_5$,  RNN$_1$ was judged to produce higher quality utterances but overall the diversity of output offered by RNN$_5$ made it the preferred system.
Even though the preference is not statistically significant, it echoes previous findings \cite{PonBarry2006,Mairesse2014} that showed that language variability by paraphrasing in dialogue systems is generally beneficial.
Lastly, RNN$_5$ was thought to be significantly better than O\&R in terms of informativeness.
This result verified our findings in Section \ref{sec:compare} that O\&R suffers from high slot error rates compared to the RNN system.

\subsection{Analysis}
\label{sec:analysis}

In order to better understand the relative contribution of each component in the RNN-based generation process, a system was built in stages
training first only the forward RNN generator, then adding the CNN reranker, and finally the whole model including the backward RNN reranker. 
Utterance candidates were reranked using Equation (\ref{eq:score}) rather than (\ref{eq:score_star}) to minimise manual intervention.
As previously, the BLEU score and slot error (ERR) were measured.

\noindent{\bf Gate} \indent The forward RNN generator was trained first with different feature gating factors $\delta$. 
Using a selection beam of 20 and selecting the top 5 utterances,
the result is shown in Figure \ref{fig:decay} for $\delta$=1 is (equivalent to not using the gate), $\delta$=0.7, and $\delta$=0 (equivalent to turning off the feature immediately its corresponding slot has been generated).
As can be seen, use of the feature gating substantially improves both BLEU score and slot error, and the best performance is achieved by setting $\delta$=0.

\noindent{\bf CNN} \indent The feature-gated forward RNN generator was then extended by adding a single convolutional-pooling layer CNN reranker.
As shown in Figure \ref{fig:slu}, evaluation was performed on both the original dataset ({\it all}) and the dataset containing only binary slots and don't care values ({\it hard}).
We found that the CNN reranker can better handle slots and values that cannot be explicitly delexicalised (1.5\% improvement on {\it hard} comparing to 1\% less on {\it all}).

\noindent{\bf Backward RNN} \indent Lastly,  the backward RNN reranker was added and trained to give the full generation model. 
The  selection beam was fixed at 100 and the $n$-best top results from which to select the output utterance was varied as $n=1$, $5$ and $10$,
trading accuracy for linguistic diversity.  In each case, the BLEU score was computed with and without the backward RNN reranker.
The results shown in Figure \ref{fig:sample} are consistent with Section \ref{sec:compare}, in which BLEU score degraded as more $n$-best utterances were chosen.
As can be seen, the backward RNN reranker provides a stable improvement no matter which value $n$ is.

\begin{figure}[h]
\centerline{\includegraphics[width=80mm]{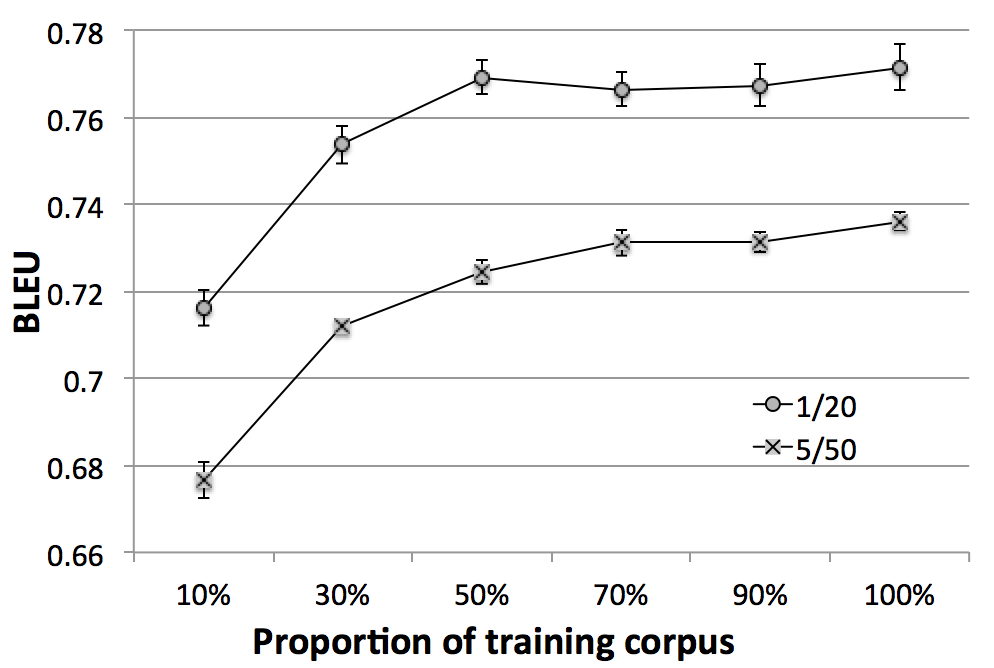}}
\caption{Networks trained with different proportion of data evaluated on two selection schemes.}
\vspace{-4 mm}
\label{fig:prop}
\end{figure}

\noindent{\bf Training corpus size} \indent 
Finally, Figure \ref{fig:prop} shows the effect of varying the size of the training corpus.
As can be seen, if only the 1-best utterance is offered to the user, then around 50\% of the data (2000 utterances) is sufficient. 
However, if the linguistic variability provided by sampling from the top-5 utterances is required, then the figure suggest that more than 4156 utterances in the current training set are required.

\section{Conclusion and Future Work}
\label{sec:conclusion}

In this paper a neural network-based natural language generator has been presented in which a forward RNN generator, a CNN reranker, and backward RNN reranker are jointly optimised to generate utterances conditioned by the required dialogue act. 
The model can be  trained on any corpus of dialogue act-utterance pairs without any semantic alignment and heavy feature engineering or handcrafting.
The RNN-based generator is compared with an $n$-gram based generator which uses similar information.  The  $n$-gram generator can achieve similar BLEU scores but it is less efficient and prone to making errors in rendering all of the information contained in the input dialogue act.

An evaluation by human judges indicated that our system can produce not only high quality but linguistically varied utterances.
The latter is particularly important in spoken dialogue systems where frequent repetition of identical output forms t.

The work reported in this paper is part of a larger programme to develop techniques for implementing open domain spoken dialogue.
A key potential advantage of neural network based language processing is the implicit use of distributed representations for words and a single compact
parameter encoding of a wide range of syntactic/semantic forms.   This suggests that it should be possible to transfer a well-trained generator of the form
proposed here to a new domain using a much smaller set of adaptation data.  This will be the focus of our future work in this area.

\section{Acknowledgements}
\vspace{-2mm}
Tsung-Hsien Wen and David Vandyke are supported by Toshiba Research Europe Ltd, Cambridge Research Laboratory.

%

\bibliographystyle{acl}
\bibliography{refs}

\end{document}